\documentclass[10pt]{ieeeaccess}
\usepackage{cite}
\usepackage{amsmath,amssymb,amsfonts}
\usepackage{algorithmic}
\usepackage{graphicx}
\usepackage{textcomp}
\usepackage{subcaption}
\usepackage[table]{xcolor}
\usepackage{multirow}
\usepackage{color} 
\usepackage{soul} 

\usepackage{bm}
\makeatletter
\AtBeginDocument{\DeclareMathVersion{bold}
\SetSymbolFont{operators}{bold}{T1}{times}{b}{n}
\SetSymbolFont{NewLetters}{bold}{T1}{times}{b}{it}
\SetMathAlphabet{\mathrm}{bold}{T1}{times}{b}{n}
\SetMathAlphabet{\mathit}{bold}{T1}{times}{b}{it}
\SetMathAlphabet{\mathbf}{bold}{T1}{times}{b}{n}
\SetMathAlphabet{\mathtt}{bold}{OT1}{pcr}{b}{n}
\SetSymbolFont{symbols}{bold}{OMS}{cmsy}{b}{n}
\renewcommand\boldmath{\@nomath\boldmath\mathversion{bold}}}
\makeatother

\def\BibTeX{{\rm B\kern-.05em{\sc i\kern-.025em b}\kern-.08em
    T\kern-.1667em\lower.7ex\hbox{E}\kern-.125emX}}
\begin{document}
\history{Date of publication xxxx 00, 0000, date of current version Jan 22, 2023.}
\doi{N/A}

\title{FaultFormer: Pretraining Transformers for
Adaptable Bearing Fault Classification}
\author{\uppercase{Anthony Zhou}\authorrefmark{1},
\uppercase{Amir Barati Farimani}\authorrefmark{1}}

\address[1]{Department of Mechanical Engineering, Carnegie Mellon University, Pittsburgh 15213, USA (e-mail: barati@cmu.edu)}

\corresp{Corresponding author: Amir Barati Farimani (e-mail: barati@cmu.edu). Courtesy appointments in Departments of Machine Learning, Chemical Engineering, and Biomedical Engineering}

\begin{abstract}
The growth of global consumption has motivated important applications of deep learning to smart manufacturing and machine health monitoring. In particular, analyzing vibration data offers great potential to extract meaningful insights into predictive maintenance by the detection of bearing faults. Deep learning can be a powerful method to predict these mechanical failures; however, they lack generalizability to new tasks or datasets and require expensive, labeled mechanical data. We address this by presenting a novel self-supervised pretraining and fine-tuning framework based on transformer models. In particular, we investigate different tokenization and data augmentation strategies to reach state-of-the-art accuracies using transformer models. Furthermore, we demonstrate self-supervised masked pretraining for vibration signals and its application to low-data regimes, task adaptation, and dataset adaptation. Pretraining is able to improve performance on scarce, unseen training samples, as well as when fine-tuning on fault classes outside of the pretraining distribution. Furthermore, pretrained transformers are shown to be able to generalize to a different dataset in a few-shot manner. This introduces a new paradigm where models can be pretrained on unlabeled data from different bearings, faults, and machinery and quickly deployed to new, data-scarce applications to suit specific manufacturing needs.
\end{abstract}


\titlepgskip=-21pt

\maketitle

\section{Introduction}
\label{sec:introduction}
The growth of emerging economies and the rapid pace of modernization in the world today are fueling an immense demand for consumer goods. Meeting this demand will require innovations in manufacturing to produce goods faster, better, and more cost-effectively. One important avenue for growth is machine health monitoring and predictive maintenance. Machine downtime is a common issue in manufacturing: most plants still operate on reactive, run-to-fail maintenance rather than predictive maintenance, resulting in 82\% of companies experiencing at least one machine failure in the last three years.\cite{ptc} The cost of this is enormous: according to a recent report, unplanned machine downtime now costs Fortune Global 500 companies 1.5 trillion dollars annually. \cite{siemens} As a result, advances in predictive maintenance can contribute to lower maintenance spending, increased capacity, and fewer wasted materials. 

Realizing these gains can come from analyzing machine sensor and vibration data. Sensors are already prevalent in factories, with nearly 50\% of manufacturers already integrating IoT sensors and devices into their processes.\cite{deloitte2023_meta} The resulting data can be both insightful and useful; vibration signals can be used to describe structural weakness, looseness in moving components, or resonance. To advance this research thrust, studies have aimed to collect data on the vibration of common failure modes, such as bearing faults, to be used to develop predictive maintenance strategies. \cite{CWRU, paderborn} 

\begin{figure*}[t!bp]
\centering
     \begin{subfigure}[b]{0.47\textwidth}
         \centering
         \raisebox{3mm}{
         \includegraphics[height=55mm]{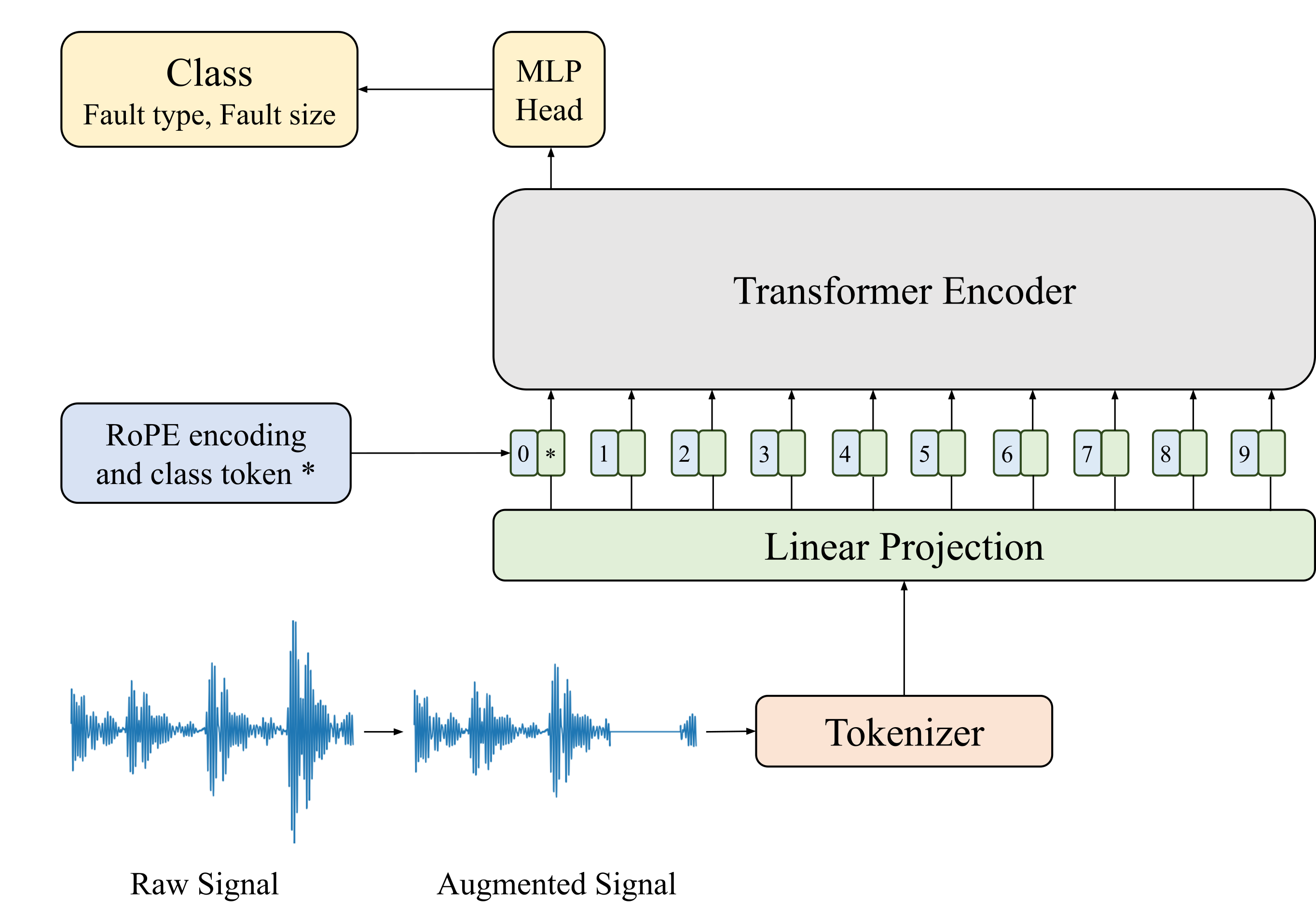}}
         \caption{End-to-end training or fine-tuning}
         \label{class}
     \end{subfigure}
     \begin{subfigure}[b]{0.49\textwidth}
         \centering
         \includegraphics[height=55mm]{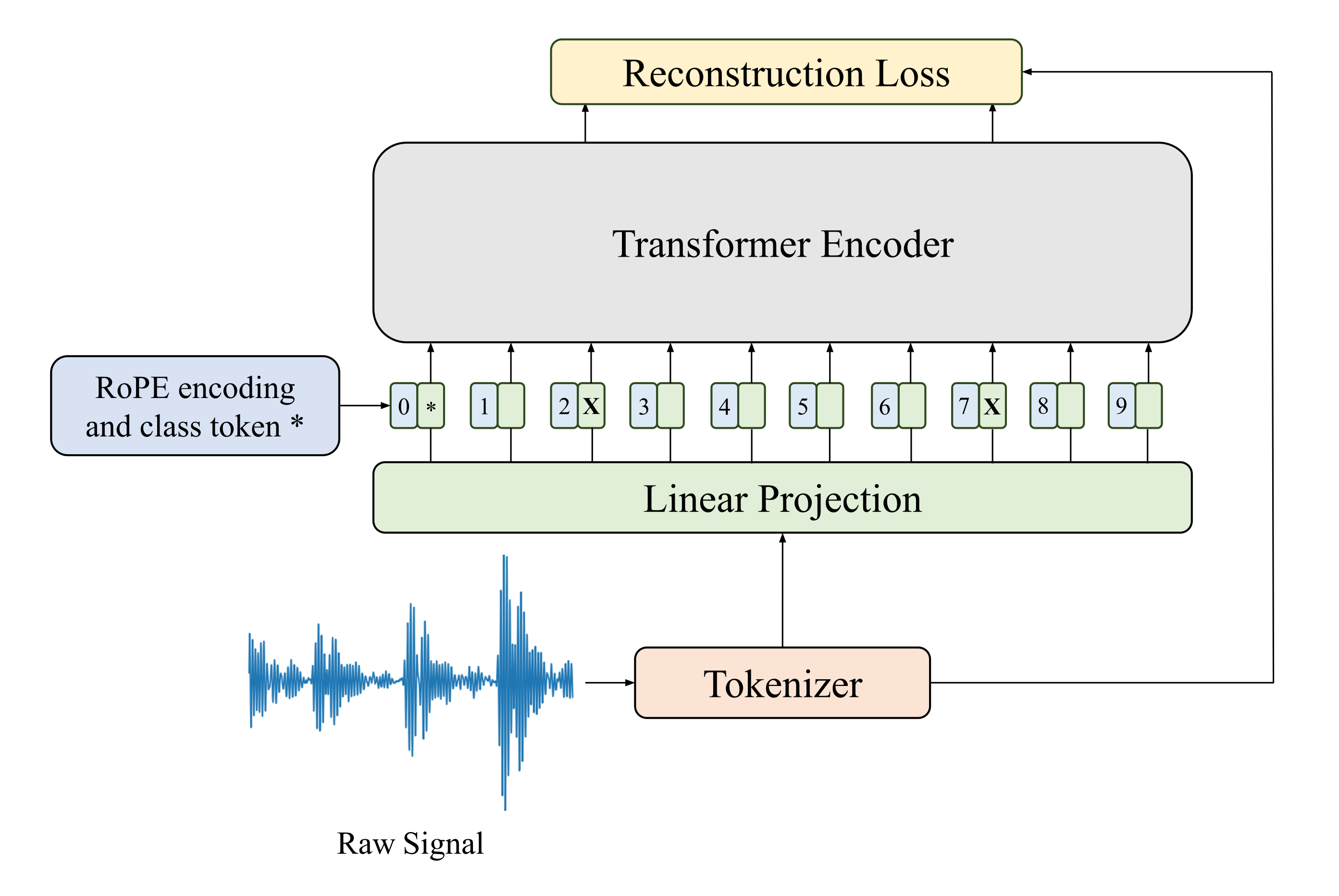}
         \caption{Masked Pretraining}
         \label{mask}
     \end{subfigure}
    \caption{\textbf{FaultFormer Architecture:} We propose using a transformer model to classify vibration data into fault classes. To aid in generalizability and improve performance, we propose pretraining transformer models on abundant, unlabeled data through self-supervised masked reconstruction.}
\label{fig:framework}
\end{figure*} 
Bearing fault detection is a rich area of study, with recent work focusing on analyzing signals from the Case Western Reserve University (CWRU) Bearing Dataset\cite{CWRU} to classify different types of bearing faults. Representative work spanning more than a decade on this problem have ranged from simple time or frequency domain approaches\cite{TANDON1994285} to newer work using deep learning.\cite{VanillaCNN} Deep learning approaches have proven to be effective and have spurred the use of CNNs, Autoencoders, RNNs, and most recently, transformers to classify signal data. \cite{VanillaCNN, Autoencoder_base, RNN1, transformer1}

Recent advances using CNNs have explored different architectures and training paradigms to achieve accuracies between 92-99\% as well as improve noise resilience and reduce both training time and data. \cite{CNNRev1, CNNRev2, CNNRev3} A survey of these approaches include vanilla CNNs\cite{VanillaCNN}, adaptive CNNs\cite{AdaptiveCNN}, multi-scale CNNs\cite{MulitscaleCNN}, and attention-guided CNNs\cite{AttentionCNN}, as well as combining time and spectral domains\cite{FusionCNN}, introducing additional featurized channels\cite{MultichannelCNN}, fusing data from multiple training samples\cite{FuseCNN}, and addressing long-tailed distributions during training \cite{longtailCNN}.

A separate research direction has proposed the use of denoising autoencoders to train a latent space which can be used as a feature extractor to fine-tune a head. Initial studies using this approach have shown success in increasing accuracy compared to neural networks without latent representations, with accuracies of 99\% compared to 70\%.\cite{Autoencoder_base} Further advances have proposed using different variants of autoencoders to better represent the latent space or reduce training data, including stacked sparse autoencoders\cite{sparse_AE}, ensemble autoencoders \cite{ensemble_AE} , and wavelet autoencoders\cite{wavelet_AE}. 

With the growing popularity of RNNs and LSTMs in deep learning, recurrent approaches to classifying vibration data have been introduced as a natural way to process sequential data. Initial studies have shown success in applying RNNs to the CWRU dataset\cite{RNN1}, and further studies have increased accuracies studies through combining RNN layers with CNN layers.\cite{RNN2} Other studies have also considered GRUs to stabilize training\cite{RNN3} and different feature extraction strategies with RNNs. \cite{RNN4}  

To further improve sequential computing, transformers were introduced as an extension of RNNs. \cite{attn_is_all_you_need}  Compared to RNNs that learn long dependencies through sequential back-propagation, transformers rely on the attention mechanism to attend to important context within sequences. This mechanism alleviates vanishing or exploding gradients in lengthy back-propagation that destroys learning signals. Modern deep learning relies heavily on the transformer to build powerful applications, as such, integrating this architecture in bearing fault classfication is an area of great interest. 

Transformer-based approaches in bearing fault classification are new, however, they are already showing promise in accuracy and interpretability. After initial work validating the use of this architecture \cite{transformer1, transformer_2023, transformer_2023_2}, different variants have been proposed to increase the robustness\cite{robust_transformer} and trustworthiness\cite{trust_transformer} of transformer predictions. However, these studies do not investigate self-supervised pretraining methods for transformers, a significant advantage of the architecture; this strategy has contributed to monumental advances in language processing and reasoning \cite{BERT, GPT}. In the context of fault detection, this would be a significant improvement on previous training paradigms, as unlabeled vibration data can be used to learn meaningful context that is applied to novel tasks. Furthermore, labeling mechanical data can be difficult due to need for real-world experiments, as a result, the ability for transformers to use unlabeled data in self-supervised learning is extremely promising. These advances would allow factories to quickly deploy pretrained deep learning models to adapt to new machinery and tasks that differ from standard datasets.

\begin{figure*}[tbp]
     \centering
     \includegraphics[width=\textwidth]{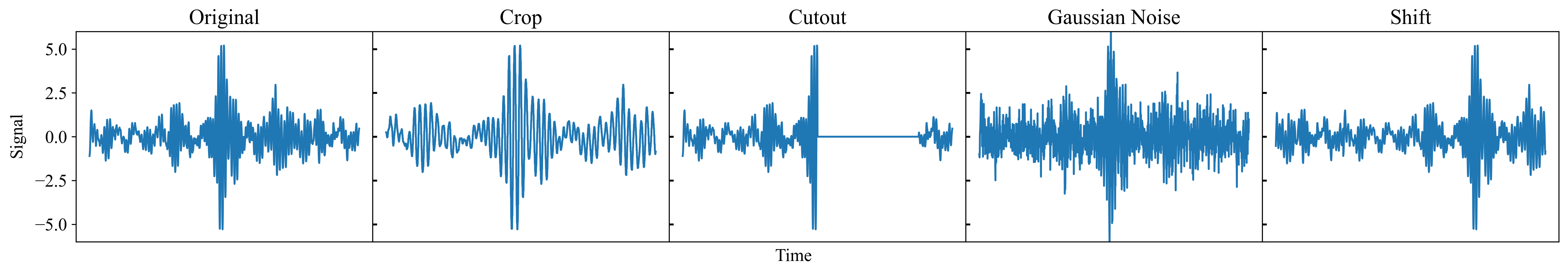}
    \caption{\textbf{Data Augmentation: }We propose random data augmentations to approximate a larger dataset for scarce real-world mechanical data. Additionally, perturbing training data can prevent overfitting.}
    \label{fig:data_augs}
\end{figure*}

A preliminary study investigating masked pretraining for transformers demonstrates its fine-tuning performance on a data-scarce, 4-way bearing classification task. \cite{pretrain_transformer} Contrastive pretraining was also considered for downstream fine-tuning on long-tailed fault distributions \cite{contrastive_transformer}, and for learning more robust latent features \cite{contrastive_2024}.  These studies verify the use of pre-training for bearing classification; however, they pretrain and fine-tune on the same classes and datasets. This data is critically sampled from the same distribution as the pretraining distribution, which this study aims to improve upon. This study demonstrates a pretrained model that adapts to new tasks through fine-tuning on data from unseen classes as well as an entirely new dataset. This scenario more closely matches real-world use cases where factory needs vary widely across different fault types and machinery, and require models that can quickly fine-tune to entirely new needs. This study presents the following contributions:

\begin{itemize}
    \item We investigate different strategies for training transformers for vibration data, including different tokenization strategies, data augmentations, and using unlabled data through masked pretraining.
    \item We validate results on adapting pretrained transformers to scarce datasets and compare different models and training sample sizes. 
    \item We demonstrate novel results on fine-tuning pretrained transformer models to predict new classes of faults as well as predicting faults from entirely new data. Fine-tuning is established as a quicker and more accurate strategy when benchmarked against different models fully trained on new tasks or data.  
\end{itemize}

\section{Methods}
\subsection{Proposed Architecture}
The proposed architecture is based on a transformer encoder that can be pretrained or fine-tuned. Raw signal data is optionally augmented then tokenized before being embedded with an MLP. A class token is prepended to the sequence of embeddings and processed with rotary positional embeddings. \cite{roformer} After passing through the transformer encoder, the class embedding is then projected to the output dimension to be used for classification. For pretraining, a mask is applied after tokenization where a percentage of tokens are either zero masked, randomly masked, or left unchanged. These tokens are linearly embedded and the resulting sequence embeddings are passed through the transformer encoding before being decoded using an MLP to evaluate a reconstruction loss. An overview of these architectures can be seen in Figure \ref{fig:framework}, and the following sections describe its individual components.

\subsubsection{Data Augmentation}
To meet the data needs of the transformer architecture and prevent overfitting, the following random data augmentations are proposed, inspired by advances in self-supervised contrastive learning.\cite{simclr} These include adding Gaussian noise, cutting out portions of the signal, cropping and resizing the signal, and shifting the signal across time, which are described in Figure \ref{fig:data_augs}. The parameters for each augmentation, such as the standard deviation of Gaussian noise or stride for a shift are uniformly sampled from a range of values, which allows us to generate a unique sample after each augmentation. During training, we can vary the amount of augmentation by defining a probability to augment training samples and leaving test samples untouched. Additional details about the parameters used can be found in Appendix A. 

\subsubsection{Signal Tokenization}
The need for a tokenizer is apparent when considering the length of input vibration signals. At thousands of data points long, directly using this data would be prohibitively expensive for transformers; furthermore, it is likely that not all data points represent useful information. As a result, three tokenization strategies are proposed: constant, CNN, and Fourier. 

The constant tokenizer simply reshapes the input signal by adding a channel dimension in addition to the time dimension. This decreases the sequence length by stacking information from multiple data points into a vector for each token. The CNN tokenizer is more sophisticated and adds a channel dimension as well as reduces the sequence length through convolving the input signal. Lastly, the Fourier tokenizer extracts the top Fourier modes of each token. The real and imaginary amplitude as well as frequency of these modes are stacked in the channel dimension. Additional details about the proposed tokenizers can also be found in Appendix B. 

\subsubsection{Transformer Encoder}
After producing tokens, a transformer encoder \cite{attn_is_all_you_need} is used to process sequential vibration data. The architecture consists of multiple stacked layers, with each layer consisting of a multi-headed self attention mechanism followed by a fully connected layer, as well as the appropriate dropout layers, residual connections, and layer normalization. 

Within each layer, the self-attention mechanism assigns an attention score to the current token based on relevance to tokens in the input sequence. These scores are calculated by a dot product between a query vector based on the current token and a key vector for each token in the sequence, and is normalized by the vector dimension $d$. The result is then passed through a softmax and used take a weighted average of the value vectors based on each input token. This operation can be represented using matrices $Q, K$ and $V$, generated from linear projections of the feature matrix $X$.
\begin{gather}
    A_n(Q, K, V) = softmax(\frac{QK^t}{\sqrt{d}}) V \\
    Q = XW^q, K = XW^k, V = XW^v
\end{gather}
This operation can be split into multiple attention heads for a given input, which allows the model to attend to different relationships for each head. Query, key and value matrices are split equally and used individually to calculate an attention score which is finally concatenated.
\begin{gather}
    Head_i = A_n((Q/h)_i, (K/h)_i, (V/h)_i) \\
    A_h(Q, K, V) = Concat(Head_1, Head_2, ..., Head_n)
\end{gather}

The output of the multi-head attention block is then passed through a feed-forward network consisting of two linear layers and a nonlinearity $\sigma$. \cite{shazeer2020glu} Finally, the output is generated by appropriate residual and layer norm operations. 
\begin{gather}
    FF(X) = \sigma(XW_1 + b_1)W_2 + b_2 \\
    X_a = LayerNorm(A_h(X) + X) \\
    X_{out} = LayerNorm(FF(X_a) + X_a) 
\end{gather} 

\begin{table}[tbp]
\centering
\tabcolsep=0.11cm
\renewcommand{\arraystretch}{1.3}
\begin{tabular}{|c|c|c|c|c|}
\hline
Experiment & Split & Dataset & \# Samples & Classes \\
\hline
\multirow{ 3}{*}{Baseline}& Pretraining & N/A & N/A & N/A \\
 & Train & CWRU & 2240 & 0-9\\
 & Test & CWRU & 560 & 0-9\\
 \hline
 \multirow{ 3}{*}{Data Scarcity}& Pretraining & CWRU$^*$ & 2000 & 0-9\\
 & Train & CWRU & 400/200/100 & 0-9\\
 & Test & CWRU & 100 & 0-9\\
 \hline
  \multirow{ 3}{*}{Task Adaptation}& Pretraining & CWRU$^*$ & 1960 & 0-2,4-5,7-8\\
 & Train & CWRU & 672 & 3,6,9 \\
 & Test & CWRU & 168 & 3,6,9 \\
 \hline
 \multirow{ 3}{*}{Dataset Adaptation}& Pretraining & CWRU$^*$ & 2800 & 0-9 \\
 & Train & Paderborn & 46400 & 0-2 \\
 & Test & Paderborn & 11600 & 0-2\\
 \hline
\end{tabular}
\caption{\textbf{Experiments and Datasets: } Datasets and splits created for different experiments. The asterisk $^*$ denotes an unlabeled dataset. Data are organized to facilitate the goals of each experiment.}
\label{tab:experiments}
\end{table}

\subsection{Experiments and Datasets}
In order to investigate transformers for bearing classification and the effects of self-supervised pretraining, a set of experiments were designed using different data splits. These experiments make use of two datasets: the Case Western Reserve University (CWRU) Bearing Dataset and the Paderborn University Dataset. The CWRU dataset is categorized into 10 classes for a total of 2800 samples, and the Paderborn dataset is categorized into 3 classes for a total of 58000 samples. Additional data about these datasets can be found in Appendix C or in the original publications. \cite{CWRU, paderborn}

A baseline is first established by training transformer models end-to-end on a standard 80/20 split of the CWRU dataset. Additionally, different tokenization and data augmentation probabilities are investigated and benchmarked against common deep learning models. Once the performance of proposed architecture is established, the effect of pretraining on improving performance in data-scarce regimes is evaluated through pretraining on an unlabeled CWRU dataset and fine-tuning on an unseen set of smaller labeled CWRU datasets. Furthermore, different models are trained from scratch in data-scarce regimes as a benchmark.

Two experiments are created to evaluate downstream performance after pretraining a transformer model. First, a transformer is pretrained from unlabeled fault data collected from a given setup and fine-tuned on new faults observed on the same setup. This corresponds to holding out data on certain classes of the CWRU dataset when pretraining, then evaluating the fine-tuned performance on new fault classes. Another experiment considers pretraining a transformer on unlabeled data from a certain setup, and fine-tuning on fault data on a different setup which could contain different motors, bearings, sensors, or defects. This is investigated by pretraining on the CWRU dataset and fine-tuning on the Paderborn dataset. A full description of the data used in each experiment can be found in Table \ref{tab:experiments}, and training parameters can be found Appendix D.

\begin{table}[tbp]
\centering
\tabcolsep=0.175cm
\renewcommand{\arraystretch}{1.3}
\begin{tabular}{|c|c|c c c c|}
\hline
  \multirow{ 2}{*}{Tokenizer}& \multirow{ 2}{*}{Model} & \multicolumn{4}{|c|}{Augmentation Probability} \\
 \cline{3-6}
 & & 0.0 & 0.3 & 0.6 & 0.9\\
  \hline
   \multirow{ 4}{*}{Fourier} & Transformer & 0.6268 & 0.6321 & 0.65 & \cellcolor[HTML]{B0B0B0}0.9984\\
   & LSTM  & 0.5286 & 0.5571 & \cellcolor[HTML]{F0F0F0} 0.7232 & \cellcolor[HTML]{D0D0D0}0.8857 \\
  & CNN & 0.3286 & 0.4482 & 0.5232 & 0.5732\\
  & MLP & 0.2571 & 0.3482 & 0.2911 & 0.5696 \\
    \hline
   \multirow{ 4}{*}{CNN} &  Transformer & \cellcolor[HTML]{F0F0F0} 0.9911 & 0.9893 & 0.9839 & 0.9821 \\
   & LSTM  & 0.9661 & \cellcolor[HTML]{D0D0D0}0.9946 & \cellcolor[HTML]{D0D0D0}0.9964 & \cellcolor[HTML]{D0D0D0}0.9946 \\
  & CNN & \cellcolor[HTML]{F0F0F0} 0.9911 & \cellcolor[HTML]{B0B0B0} 0.9982 & \cellcolor[HTML]{F0F0F0} 0.9911 & \cellcolor[HTML]{B0B0B0}0.9982 \\
  & MLP & 0.9054 & \cellcolor[HTML]{B0B0B0} 0.9982 & \cellcolor[HTML]{D0D0D0}0.9964 & \cellcolor[HTML]{D0D0D0}0.9964 \\
    \hline
    \multirow{ 4}{*}{Constant} & Transformer & 0.9107 & 0.9321 & 0.925 & 0.9089 \\
   & LSTM  & 0.9929 & \cellcolor[HTML]{F0F0F0} 0.9946 & \cellcolor[HTML]{D0D0D0} 0.9964 & \cellcolor[HTML]{B0B0B0} 0.9982 \\
  & CNN & \cellcolor[HTML]{D0D0D0} 0.9964 & 0.9911 & 0.9929 & 0.9911 \\
  & MLP & 0.8946 & 0.9679  & 0.9875 & \cellcolor[HTML]{F0F0F0} 0.9946 \\
    \hline
\end{tabular}
\caption{\textbf{Baseline Results: } Test accuracies of different models under various tokenizer and augmentation parameters. Top 3 accuracies are highlighted, using darker colors for higher values.}
\label{tab:results}
\end{table}

\section{Results and Discussion}

\subsection{Baseline Results}
The effects of data augmentation and different tokenizers were investigated by training different transformer variations on the CWRU dataset. Furthermore, LSTM, CNN, and MLP models were also trained as a benchmark. Test accuracies can be found in Table \ref{tab:results}. 

In general, high-performing models make use of frequent data augmentations. We observe that transformer models can perform just as well as benchmark models, and outperform other models when analyzing data in the Fourier domain. These results suggest that transformer models have similar capacity of state-of-the-art models in classifying vibration data. 

The performance of transformer models in the Fourier domain can be explained through considering the use of data augmentations. We observed that LSTM, CNN, and MLP models severely overfit to the training set, quickly reaching accuracies of 100\% yet having low test accuracies. This could be because using a Fourier basis allows the data to be represented by coefficients that describe global trends in the vibration signal, allowing for compact representations of trends within inputs. However, this reduces the amount of available data, which could contribute to overfitting. As a result, data augmentation can be used to create more diverse and unique training samples to increase performance. Using data augmentations makes the training task significantly more difficult as well as adds more data to the training set, as augmentations in the time domain substantially affect the Fourier coefficients. Therefore, the expressive power of the transformer architecture can be utilized through learning a difficult, data-rich task and evaluated on a simpler, yet unseen data. 

Another relevant insight is the competitive performance of simpler models when using data augmentations. Even simple MLP architectures can reach accuracies of 99\%, a model that is not conventionally considered in literature. Additionally, we see that transformer models struggle without a tokenization strategy. Naively reshaping input sequences is detrimental to transformer models since the attention mechanism can only resolve relationships between and not within tokens. By simply grouping data within tokens without extracting features, the model loses the ability to attend to meaningful information within tokens. 

\begin{figure}[tbp]
     \centering
     \includegraphics[width=0.48\textwidth]{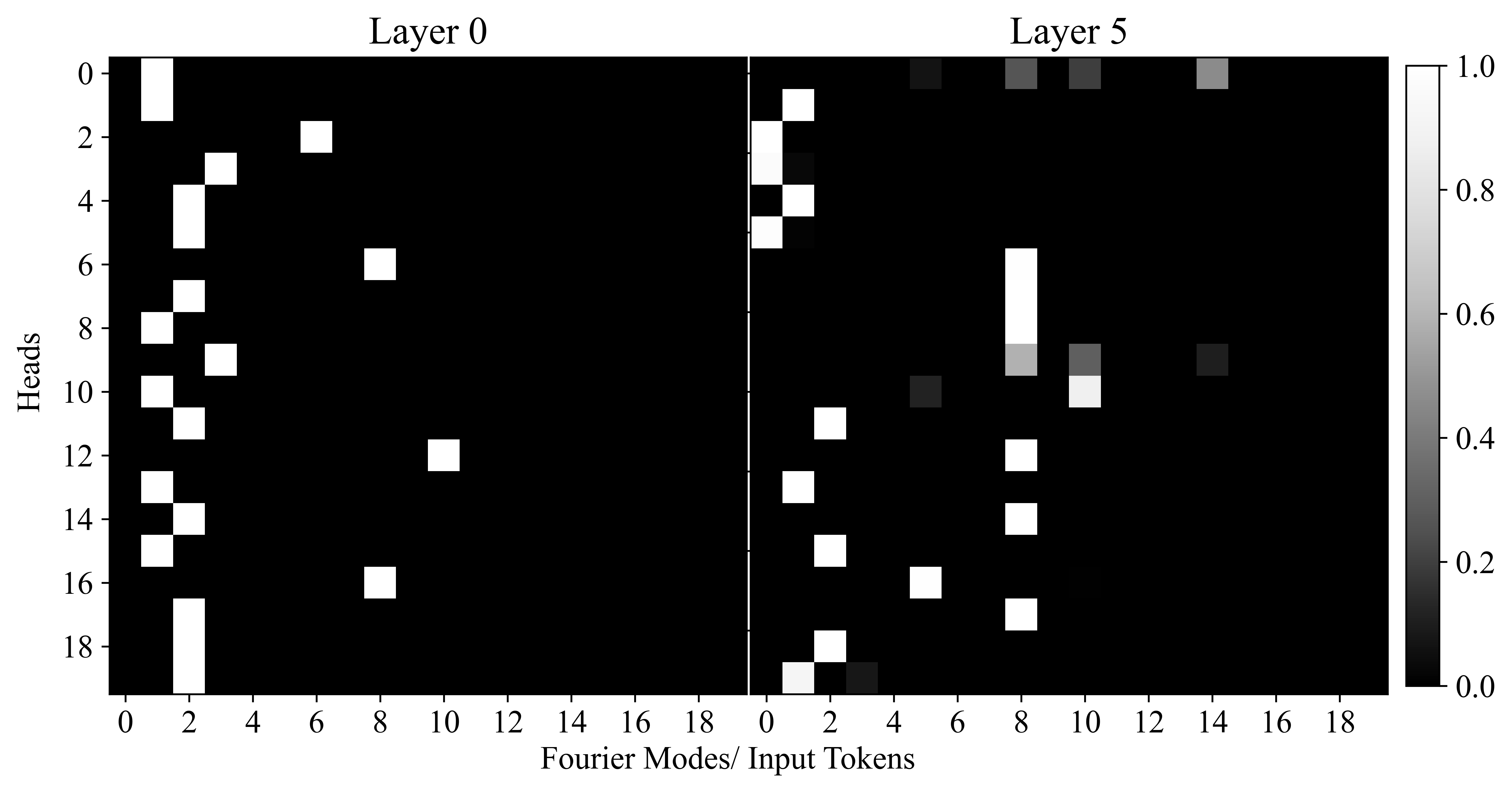}
    \caption{\textbf{Attention to Fourier Modes: } Attention scores of input tokens visualized for each transformer head using Fourier tokenization. At shallow layers transformer heads attend to single, dominant Fourier modes in the input sequence. At deeper layers, transformer heads attend to multiple as well as smaller Fourier modes. This corresponds to learning both global and fine relationships in input signals. }
    \label{fig:attention}
\end{figure}

\begin{figure}[t!bp]
     \centering
     \includegraphics[width=0.45\textwidth]{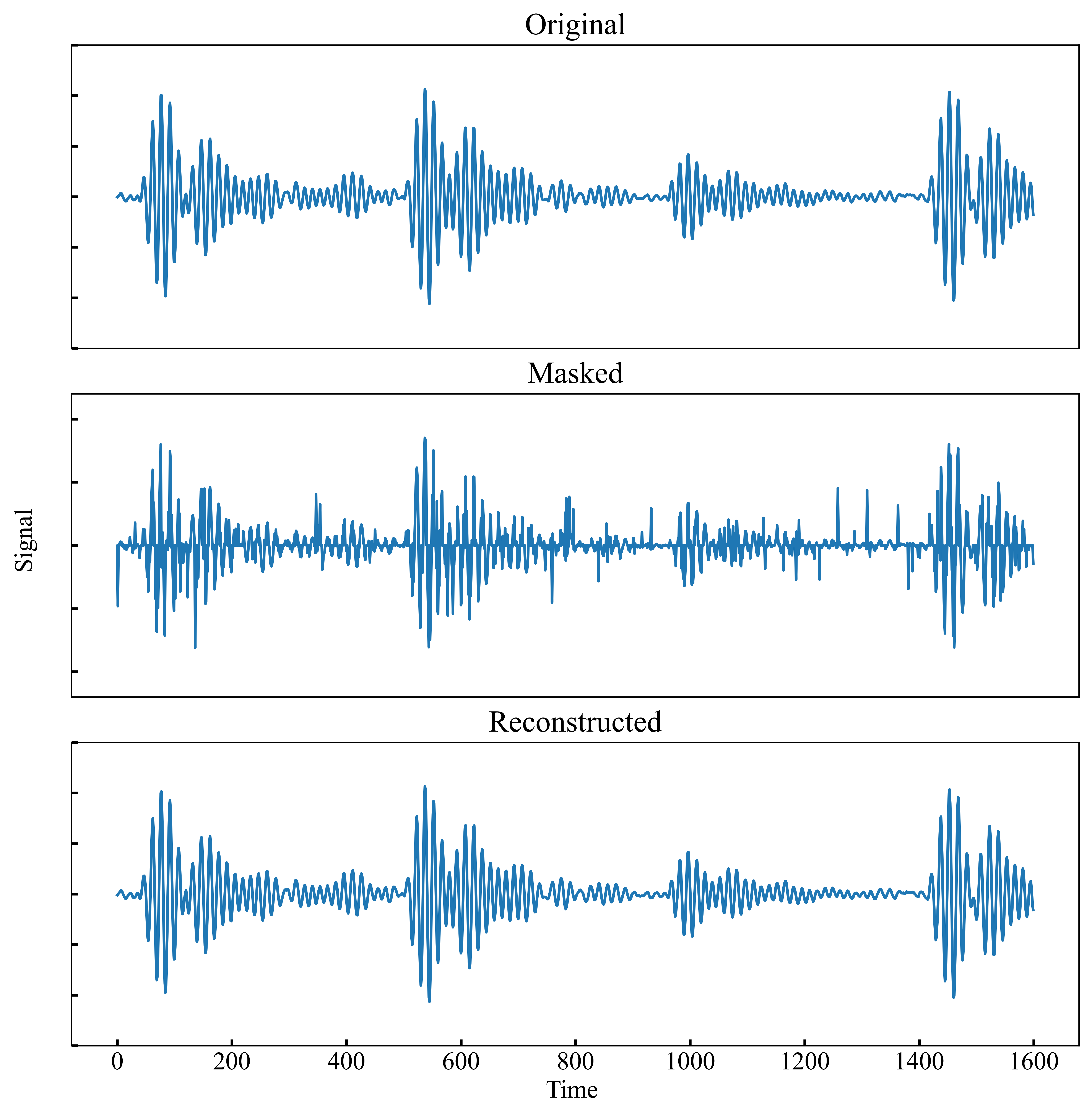}
    \caption{\textbf{Masked Pretraining: }An example signal that is masked and reconstructed with a transformer encoder. The model is able to reconstruct signals without noticeable error.}
    \label{fig:masked}
\end{figure}

\subsection{Visualizing Results}

In order to further investigate the transformer's performance in the Fourier basis, the attention scores for different layers are visualized in Figure \ref{fig:attention}. In particular, attention scores were calculated between the class token and all other tokens for each attention head in a layer. In the first layer, the heads attend to a single token/mode, and many heads attend to the same token. Furthermore, the heads attend to the largest Fourier modes, which have the most impact on the signal. However, at deeper layers, the heads attend to a combination of tokens/modes, and each head differentiates to attend to different sets of tokens. Smaller Fourier modes also receive attention, suggesting that the model is capable of attending to both coarse and fine details in input signals. These visualizations suggest that the transformer architecture is indeed capable of capturing complex, multiscale features in input sequences and aggregating the information by encoding the class token. 

\begin{table}[t!bp]
\centering
\tabcolsep=0.175cm
\renewcommand{\arraystretch}{1.3}
\begin{tabular}{|c|c c c|}
\hline
    \multirow{ 2}{*}{Model} & \multicolumn{3}{|c|}{Training Sample Size} \\
 \cline{2-4}
  & 100 & 200 & 400 \\
  \hline
     Transformer-PT & \cellcolor[HTML]{B0B0B0} 0.9327 & \cellcolor[HTML]{B0B0B0} 0.9615 & \cellcolor[HTML]{B0B0B0} 0.9615 \\
     Transformer & 0.8077 & 0.8173 & 0.8462 \\
     LSTM & 0.8654 & 0.9519 & \cellcolor[HTML]{B0B0B0} 0.9615\\
   CNN & 0.9135 & \cellcolor[HTML]{B0B0B0} 0.9615 & \cellcolor[HTML]{B0B0B0} 0.9615 \\
   MLP & 0.7596 & 0.8654 & 0.9038 \\
  \hline
\end{tabular}
\caption{\textbf{Data Scarcity: }Test accuracies of different models given various training sample sizes. The highest accuracy for each sample size is highlighted. Pretraining increases performance in low-data regimes.}
\label{tab:scarce}
\end{table}

\subsection{Masked Pretraining}
After validating the ability of transformer models to classify bearing fault data and reach state-of-the-art accuracies, we investigate the use of masked pretraining on unlabeled vibration data. In this study, we follow a strategy used in language modeling in which we mask 50\% of tokens; of those tokens, 70\% are set to zero, 20\% are replaced to a random value in the signal, and 10\% are left untouched. Aggressively masking tokens is helpful due to learning more expressive latent representations, as well as signal data being easier to reconstruct than language tokens. Qualitative results are presented in Figure \ref{fig:masked}, and show that the transformer encoder is able to learn to nearly perfectly reconstruct masked signals.

\begin{figure}[tbp]
  \centering
  \begin{minipage}[b]{1.0\linewidth}
    \centering
    \label{fig:sub-a}
    \includegraphics[width=0.85\textwidth]{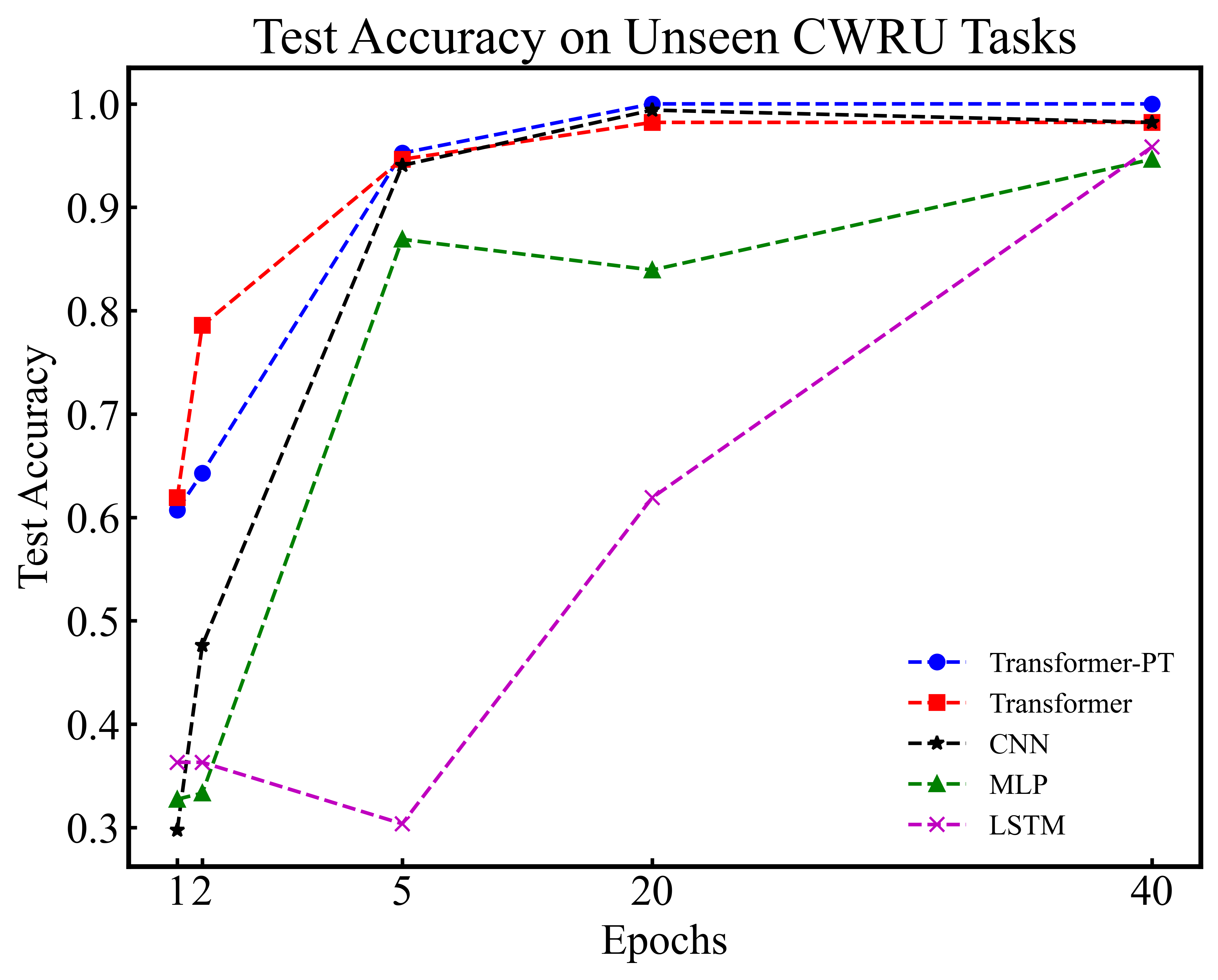}
    \qquad
\tabcolsep=0.175cm
\renewcommand{\arraystretch}{1.35}
\fontsize{8pt}{8pt}\selectfont
    \begin{tabular}{|c|c c c c c|}
        \hline
            \multirow{ 2}{*}{Model} & \multicolumn{5}{|c|}{Epochs} \\
         \cline{2-6}
          & 1 & 2 & 5 & 20 & 40\\
          \hline
             Transformer-PT & 0.6071 & 0.6429 & \cellcolor[HTML]{B0B0B0} 0.9524 & \cellcolor[HTML]{B0B0B0} 1.000 & \cellcolor[HTML]{B0B0B0} 1.000\\
             Transformer & \cellcolor[HTML]{B0B0B0} 0.6190 & \cellcolor[HTML]{B0B0B0} 0.7857 & 0.9464 & 0.9821 & 0.9821 \\
            LSTM & 0.3631 & 0.3631 & 0.3036 & 0.6190 & 0.9583\\
           CNN & 0.2976 & 0.4762 & 0.9405 & 0.9940 & 0.9821 \\
           MLP & 0.3274  & 0.3333 & 0.8690 & 0.8393 & 0.9464 \\
          \hline
\end{tabular}
  \end{minipage}
  \caption{\textbf{Task Adaptation: }Test accuracies of different models at various epochs when trained on a subset of CWRU faults. The highest accuracy for each epoch is highlighted. Pretrained transformer models can achieve higher accuracies more quickly.}
  \label{fig:taskadapt}
\end{figure}

\subsection{Data Scarcity}
After pretraining a transformer model, we investigate its performance in low-data regimes. In particular, we consider a 10-way classification task when provided only 100, 200 and 400 training samples, and results are presented in Table \ref{tab:scarce}.

We observe that self-supervised pretraining increases performance when fine-tuning on small amounts of unseen data. As the number of training samples increases, the benchmark models converge to the same performance; however, pretrained models can reach optimal performance with fewer samples. Interestingly, we observe that transformers without pretraining do not seem to perform well, possibly since transformer models are known to need large quantities of data due to a lack of inductive bias. These results suggest that pretraining is essential for transformers to perform in low-data regimes. 

\subsection{Task Adaptation}
In order to evaluate the effect of pretraining on classifying data from unseen fault classes, we pretrain and fine-tune on data split by fault class. After pretraining on unlabeled data from healthy, inner, outer, and ball fault classes, we fine-tune on inner, outer, and ball fault classes with fault sizes larger than the pretraining dataset to simulate adaptation to unseen faults. The results are shown in Figure \ref{fig:taskadapt}.

The results show that pretraining is able to achieve better final performance as well as quickly adapt to new, unseen tasks. Although a transformer architecture trained end-to-end initially outperforms the pretrained transformer, this could be because the pretrained architecture is initialized with a random head. During fine-tuning, the head is trained for a few epochs before utilizing relevant context from the pretrained encoder. We note that all models tend to converge to high accuracies, however the benefit of pretraining is in its ability to reach high accuracies more quickly and with less computation. This suggests that pretrained models can be quickly adapted to new bearing faults through learning meaningful context from other faults. 

\begin{figure}[tbp]
  \centering
  \begin{minipage}[b]{1.0\linewidth}
    \centering
    \label{fig:sub-a}
    \includegraphics[width=0.85\textwidth]{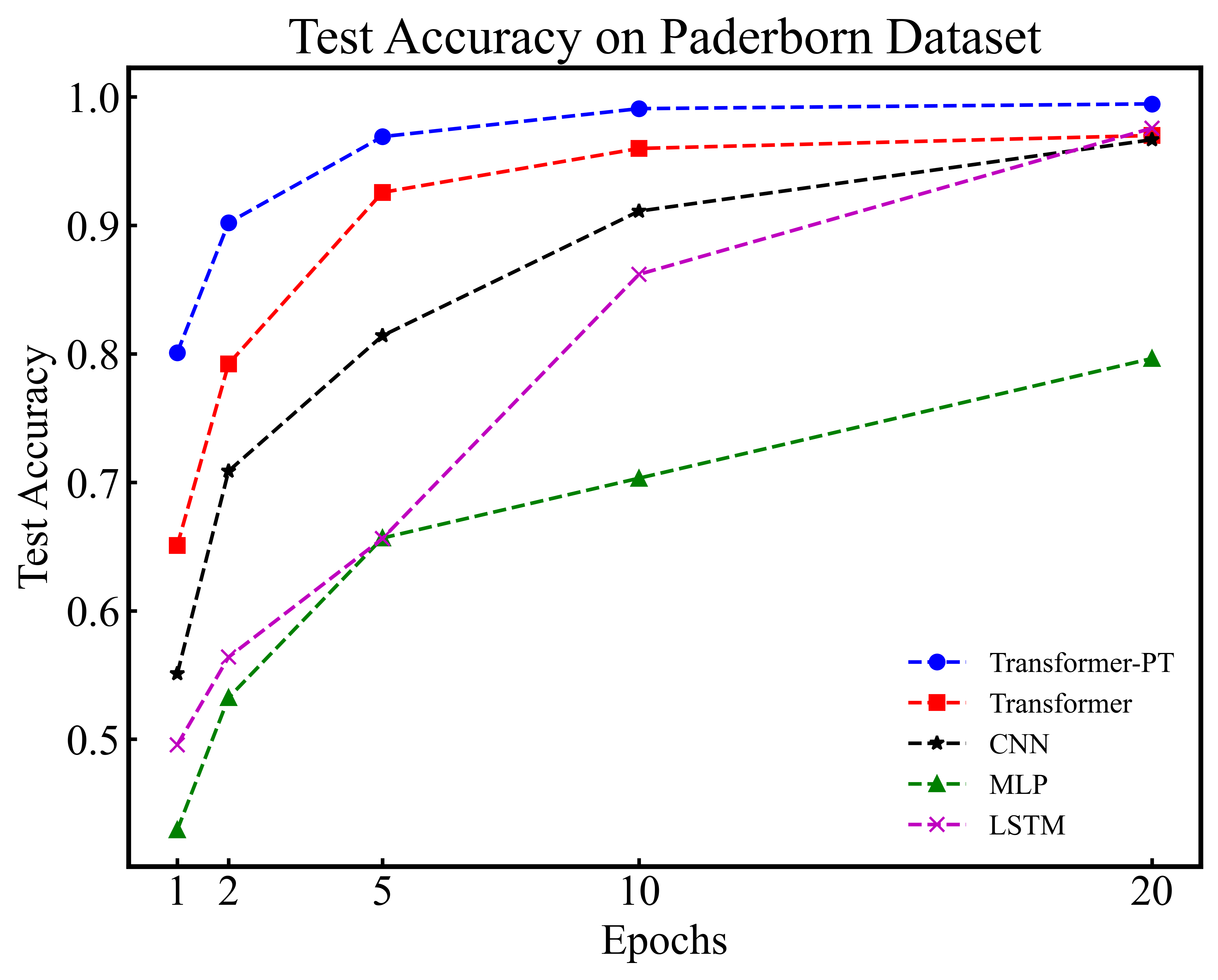}
    \qquad

    \tabcolsep=0.175cm
    \renewcommand{\arraystretch}{1.35}
    \fontsize{8pt}{8pt}\selectfont
        \begin{tabular}{|c|c c c c c|}
            \hline
                \multirow{ 2}{*}{Model} & \multicolumn{5}{|c|}{Epochs} \\
             \cline{2-6}
              & 1 & 2 & 5 & 10 & 20\\
              \hline
             Transformer-PT & \cellcolor[HTML]{B0B0B0} 0.8010 & \cellcolor[HTML]{B0B0B0} 0.9021 & \cellcolor[HTML]{B0B0B0} 0.9691 & \cellcolor[HTML]{B0B0B0} 0.9909 & \cellcolor[HTML]{B0B0B0} 0.9946 \\
             Transformer & 0.6509 &  0.7922 & 0.9257 & 0.9599 & 0.9700\\
             LSTM & 0.4956 & 0.5639 & 0.6562 & 0.8619 & 0.9757 \\
           CNN & 0.5509 & 0.7088 & 0.8141 & 0.9111 & 0.9667\\
           MLP & 0.4293 & 0.5323 & 0.6566 & 0.7032 & 0.7964\\
              \hline
    \end{tabular}
  \end{minipage}
  \caption{\textbf{Dataset Adaptation: }Test accuracies of different models at different epochs when training on the Paderborn dataset. The highest accuracy for each epoch is highlighted. Pretrained transformer models show higher accuracies across all epochs.}
  \label{fig:dataadapt}
\end{figure}

\subsection{Dataset Adaptation}
To evaluate the ability of pretrained models to generalize to entirely new data collected from unseen bearings, motors, or sensors, we pretrain a transformer on unlabeled data from the CWRU dataset and fine-tune on data from the Paderborn dataset. These results can be seen in Figure \ref{fig:dataadapt}.

These results are promising, since we observe that pretrained models can generalize to new datasets in a few-shot manner. After two epochs, pretrained transformer models can achieve greater than 90\% accuracy, which is 5 times faster than the CNN benchmark. Furthermore, pretraining is able to achieve a higher final accuracy. These results suggest that the context learned through masked pretraining can generalize between datasets, opening up the possibility to pretrain on unlabeled data across different machinery, faults, and operating conditions. This could allow a single model to perform across a variety of scenarios, as well as fine-tune to entirely new situations without extensive labeled data. 

\section{Discussion}
Recent advances in deep learning have focused on pretraining large-scale models, both to quickly fine-tune to a variety of tasks as well as benefit from shared learning on large, heterogeneous datasets. Pivotal contributions in the transformer architecture\cite{attn_is_all_you_need} and self-supervised pretraining\cite{GPT} have enabled this paradigm shift in the natural language and vision domains. Translating these proven benefits to mechanical and manufacturing domains offers great potential, which is explored in this work. 

Similar to broader deep learning studies, self-supervised pretraining for mechanical applications allows models to benefit from a significantly larger unlabeled dataset, especially when labeled, real-world data is scarce. Furthermore, self-supervised training can still learn powerful latent representations through proxy tasks; several theoretical works have investigated this phenomenon, including the ability of models to learn emergent behaviors or approximate reasoning when using only unlabeled data \cite{Radford2019LanguageMA, DBLP:journals/corr/abs-2005-14165}.

In the context of machine health monitoring, it is likely that different mechanical faults can share common features that allow self-supervised models to leverage past knowledge, even when encountering entirely new data. Furthermore, we demonstrate that even simple MLP models can learn to classify bearing faults; however, the challenge lies in the time, computing, dataset collection/labeling, and training pipeline to reach this optimum. Fine-tuning a simple model on top of a pretrained model effectively circumvents these challenges, allowing models to leverage a powerful initialization that is conditioned on previous time, compute, and data resources to quickly achieve optimal performance. 

\section{Conclusion}
In this study a framework for pretraining and fine-tuning transformer models for bearing fault classification is introduced. The framework uses the transformer architecture as an encoder to process vibration data in parallel and classifies the signal through encoding an output token. Different tokenization and data augmentation strategies are introduced to improve the performance of the transformer and benchmark models, demonstrating that transformers can match current deep learning approaches. Investigating the mechanisms of the transformer encoder shows that the class token is able to learn diverse dependencies on tokens in the input sequence. Attention heads are able to focus on different Fourier modes at different layers, thus allowing the class token to capture both coarse and fine details in the input sequence. 

To utilize the full potential of the transformer architecture, it is pretrained using masked self-supervised learning and evaluated with three experiments. Pretrained models are able to outperform models in low-data regimes due to learning meaningful context from separate, unlabeled data. These capabilities are further expanded upon by fine-tuning on unseen fault classes. Pretrained models can quickly extrapolate to fault classes outside of the pretraining distribution and achieve higher accuracies than end-to-end training. Furthermore, pretrained models can adapt to unseen datasets and generalize to new fault, machinery, or operating conditions in a few-shot manner. 

Shifting the paradigm of bearing classification from end-to-end training to pretraining/fine-tuning offers the potential for more flexible models in practice. Rather than retraining models for specific tasks, pretrained models can benefit from self-supervised learning on abundant, heterogeneous, unlabeled data and quickly adapt to an unseen task. Despite these benefits, we note that effective pretraining requires an extremely large dataset, albeit unlabeled. Additionally, in critical maintenance applications, deep learning predictions can be difficult to understand or quantify their uncertainty. Furthermore, the distribution of unlabeled data is likely skewed to be heavily in favor of healthy bearings due to it being the typical operating condition, as a result methods for weighting faulty samples could be of interest for future work. Additional future work could also scale up to larger transformer models and more diverse unlabeled datasets to fully realize its benefits. Different pretraining methods for unlabeled mechanical data beyond masked pretraining can also be explored; moreover, multiple modes of data, such as video or audio monitoring data can be included. In general, we believe that there is great potential in foundational models for mechanical data, which can quickly perform diverse, real-world tasks on the manufacturing floor. 

\appendices
\section{\break Data Augmentation}
The strategy used to augment data is to sample a probability uniformly on $[0, 1]$ and if it is less than the data augmentation probability, we choose between eight possibilities with equal probability:

\begin{itemize}
    \item Gaussian Noise: Gaussian noise with a standard deviation sampled uniformly from $[0, 0.05]$ is added to the signal.
    \item Shift: The signal is shifted by a timestep sampled uniformly from $[-\frac{l}{2}, \frac{l}{2}]$, where $l$ is the signal length.
    \item Cutout: A window of the signal is zeroed out, with the window length sampled uniformly from $[100, 500]$. The window starts at a point sampled uniformly from $[0, l-l_w]$, where $l_w$ is the window length.
    \item Crop: A window of length $\frac{l}{2}$ is chosen from the signal starting at a random point sampled uniformly on $[0, \frac{l}{2}]$. The window is then upsampled using linear interpolation to the original signal length.
    \item Cutout + Shift: The shift operation is performed followed by the cutout operation.
    \item Cutout + Gaussian Noise: The gaussian noise operation is performed followed by the cutout operation.
    \item Crop + Shift: The shift operation is performed followed by the crop operation.
    \item Crop + Gaussian Noise: The gaussian noise operation is performed followed by the crop operation.
\end{itemize}

\section{\break Signal Tokenization}
\subsection{Constant Tokenizer}
The constant tokenizer reshapes the input sequence, where $d$ is the channel dimension. 
\begin{equation*}
    (batch\_size, seq\_length, 1) \rightarrow (batch\_size, seq\_length/d, d)
\end{equation*}

\subsection{CNN Tokenizer}
The CNN tokenizer passes the input sequence through a set of 1D convolutional layers described in Table \ref{tab:cnntokenizer}.

\begin{table}[htbp]
	\centering
	\caption{CNN Tokenizer Layers}
	\label{tab:cnntokenizer}
	\begin{tabular}{lccc}
	\hline
	Layer & Output Shape & Kernel Size & Stride\\
	\hline
    Input & (B, 1, L) & - & - \\
    Conv1D & (B, 4, L/2) & 4 & 2 \\
    GELU & - & - & - \\
    Conv1D & (B, 8, L/4) & 4 & 2 \\
	\hline
	\end{tabular}
\end{table}

\subsection{Fourier Tokenizer}
The Fourier tokenizer is implemented by taking the FFT of an input signal and extracting the top 40 modes. The real and imaginary amplitudes, as well as the frequency is stacked in the channel dimension. 

\begin{equation*}
    (batch\_size, seq\_length, 1) \rightarrow (batch\_size, 40, 3)
\end{equation*}

\section{\break Dataset Details}
\subsection{Case Western Reserve University Dataset}
The CWRU dataset is generated from a test rig consisting of a 2hp electric motor driving a shaft with a torque transducer, encoder, and dynamometer. The test rig also consists of bearings mounted at the drive end and fan end of the motor, which are varied based on the test case. Faults in the test bearing were introduced at different locations using electronic discharge machining, with fault diameters ranging from .18 to .53 millimeters to form a total of 10 test cases, described in Table \ref{tab:cwru}. To collect data, motor loads were varied between 0 and 3 hp, and vibrations were measured using accelerometers attached to the motor housing at 12 kHz and 48 kHz. 

\begin{table}
\centering
\begin{tabular}{|c c c c|}
\hline
 Fault Type & Fault Size(mm) & No. Samples & Class Label \\ [0.5ex] 
 \hline 
 Normal & N/A & 280 & 0 \\ 
 \hline
 Ball Fault & .18 & 280 & 1 \\
 \hline
 Ball Fault & .36 & 280 & 2 \\
 \hline
 Ball Fault & .53 & 280 & 3 \\
 \hline
 Inner Race Fault & .18 & 280 & 4 \\
 \hline
  Inner Race Fault & .36 & 280 & 5 \\
 \hline
  Inner Race Fault & .53 & 280 & 6 \\
 \hline
  Outer Race Fault & .18 & 280 & 7 \\
 \hline
  Outer Race Fault & .36 & 280 & 8 \\
 \hline
  Outer Race Fault & .53 & 280 & 9 \\
 \hline
\end{tabular}
\caption{Description of classes in the CWRU dataset}
\label{tab:cwru}
\end{table}

The data chosen for the analysis was collected at 48 kHz for the load condition of 2hp and 1750 rpm. Since the motion is periodic, the data is preprocessed such that each sample contains data for a single revolution. At 1750 rpm and 48 kHz, this amounts to around 1670 data points per sample, which is reduced to 1600 points after truncating the first and last 35 points. As such, the 467600 provided data points for each fault class is split into 280 samples each containing 1600 points. With 10 classes, the preprocessed dataset consists of 2800 samples, with each class containing 280 samples. Further details of the test setup and data collection can be found at the CWRU Bearing Dataset website.

\subsection{Paderborn University Dataset}

To evaluate the ability of the FaultFormer model to make predictions across different datasets, the Paderborn University dataset was also considered. The Paderborn dataset consists of 32 bearings which are split into 6 healthy bearings, 12 artificially damaged bearings, and 14 naturally damaged bearings. These bearings are split into three classes based on their faults: healthy, inner race fault, or outer race fault. Three bearings with both inner and outer race faults were omitted in this analysis. The loading condition and sampling rate chosen were 1500 rpm, .7 Nm of torque, and 1000 N of radial force.

Each bearing is used 20 times for each loading condition, resulting in a total of 580 signals. Each signal is generated from 4 seconds of sampling at 64 kHz, which is split into 100 different samples, resulting 58000 samples to be classified into three classes. 

\section{\break Implementation Details}

\subsection{Transformer Encoder: Finetuning}
The transformer encoder was implemented using the x-transformers repository (https://github.com/lucidrains/x-transformers). This allowed easy use of RoPE, flash attention, and GLU activations on the feed-forward networks. Furthermore, an AdamW optimizer and one-cycle scheduler was used to warmup the learning rate during training. The full hyperparameters can be found in Table \ref{tab:hyperparams}.

\begin{table}[htbp]
	\centering
	\caption{Hyperparameters during fine-tuning}
	\label{tab:hyperparams}
	\begin{tabular}{lc}
	\hline
	Hyperparameter & Value \\
	\hline
	Batch size & 16 \\
    Epochs & 1000 \\
	Input dimension & variable \\
	Model dimension & 256 \\
	Number of attention heads & 32 \\
    Number of layers & 4 \\
    Dropout & 0.3 \\
	Warmup steps & 100 \\
    Minimum Learning Rate & 1e-4\\
    Maximum Learning Rate & 1e-3\\
    Beta 1 & 0.9 \\
    Beta 2 & 0.98 \\
	\hline
	\end{tabular}
\end{table}

\subsection{Transformer Encoder: Pretraining}
During pretraining, the architecture is slightly modified due to the lack of a decoder and use of a mask. This results in slightly different hyperparameters, shown in Table \ref{tab:hyperparams_pretrain}.
\begin{table}[htbp]
	\centering
	\caption{Hyperparameters during Pretraining}
	\label{tab:hyperparams_pretrain}
	\begin{tabular}{lc}
	\hline
    Hyperparameter & Value \\
	\hline
	Batch size & 16 \\
    Epochs & 1000 \\
	Input dimension & variable \\
	Model dimension & 256 \\
	Number of attention heads & 32 \\
    Number of layers & 4 \\
    Dropout & 0.3 \\
	Warmup steps & 100 \\
    Minimum Learning Rate & 1e-4\\
    Maximum Learning Rate & 1e-3\\
    Beta 1 & 0.9 \\
    Beta 2 & 0.98 \\
    Mask probability & 0.5 \\
    Random mask probability & 0.2 \\
    Replace probability & 0.9 \\
	\hline
	\end{tabular}
\end{table}

\subsection{CNN}
The CNN architecture is described below in Table \ref{tab:cnn}.

\begin{table}[htbp]
	\centering
	\caption{CNN Architecture}
	\label{tab:cnn}
	\begin{tabular}{lccc}
	\hline
	Layer & Output Shape & Kernel Size & Stride\\
	\hline
    Input & (B, 8, L) & - & - \\
    Conv1D & (B, 32, L/2) & 10 & 2 \\
    GELU & - & - & - \\
    Conv1D & (B, 256, L/2) & 5 & 1 \\
    GELU & - & - & -  \\
    Conv1D & (B, 512, L/2) & 3 & 1 \\
    GELU & - & - & -\\
    Conv1D & (B, 256, L/2) & 3 & 1 \\
    GELU & - & - & - \\
    Conv1D & (B, 256, L/4) & 3 & 2 \\
    AdaptiveAvgPool & (B, 256, 8) & -& - \\
    Flatten & (B, 2048) & - & - \\
    Dropout & - & - & - \\
    Linear & (B, 512) & - & - \\
    GELU & - & - & - \\
    Dropout & -& -& -\\
    Linear & (B, 526) & -& -\\
    GELU & -& -& -\\
    Linear & (B, 10) & -& -\\
	\hline
	\end{tabular}
\end{table}

\subsection{MLP}
The MLP architecture is described below in Table \ref{tab:mlp}. Adaptive average pooling is used in order to facilitate different input sequence lengths, as the sequence is pooled along its length dimension to the model dimension.

\begin{table}[htbp]
	\centering
	\caption{MLP Architecture}
	\label{tab:mlp}
	\begin{tabular}{lc}
	\hline
	Layer & Output Shape \\
	\hline
    Input & (B, 8, L)  \\
    AdaptiveAvgPool & (B, 256, 8)  \\
    Flatten & (B, 2048) \\
    Dropout & -  \\
    Linear & (B, 1024)  \\
    GELU & - \\
    Dropout & -\\
    Linear & (B, 1024) \\
    GELU & - \\
    Linear & (B, 512) \\
    GELU & -\\
    Linear & (B, 256) \\
    GELU & -\\
    Linear & (B, 10) \\
	\hline
	\end{tabular}
\end{table}

\section*{Acknowledgment}
The authors would like to thank Cooper Lorsung for insightful discussions regarding the application of transformers to scientific and engineering problems. 

\bibliographystyle{unsrt}
\bibliography{main}

\begin{IEEEbiography}[{\includegraphics[width=1in,height=1.25in,clip,keepaspectratio]{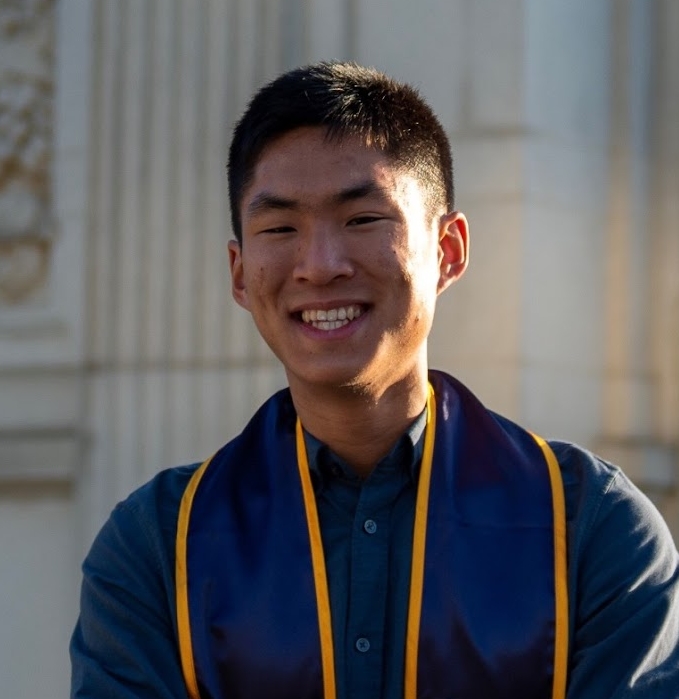}}]{Anthony Y. Zhou} received a B.S. degree in Mechanical Engineering from the University of California-Berkeley, in 2023, and is currently pursuing a Ph.D. degree in mechanical engineering at Carnegie Mellon University. He currently works in the Mechanical and AI Lab (MAIL), advised by Dr. Amir Barati Farimani.His research interests lie at the intersection of mechanical engineering and machine learning, with applications to surrogate models for partial differential equations.
\end{IEEEbiography}

\begin{IEEEbiography}[{\includegraphics[width=1in,height=1.25in,clip,keepaspectratio]{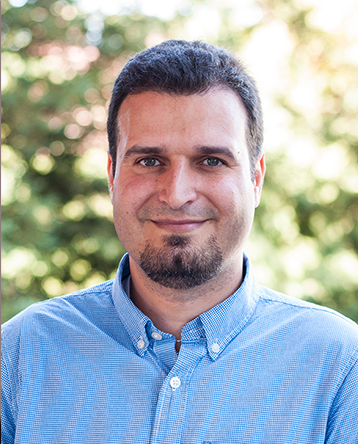}}]{Amir Barati Farimani} received his Ph.D. in 2015 in mechanical science and engineering from the University of Illinois at Urbana-Champaign. His Ph.D. thesis was titled “Detecting and Sensing Biological Molecules using Nanopores.” He extensively used atomistic simulations to shed light on the DNA sensing and detection physics of biological and solid state nanopores. Right after that, he joined Professor Vijay Pande’s lab at Stanford. During his post-doc, he combined machine learning and molecular dynamics to elucidate the conformational changes of G-Protein Coupled Receptors (GPCRs). He specifically was focused on Mu-Opioid Receptors to elucidate their free energy landscape and their activation mechanism and pathway. Amir Barati Farimani’s lab, the Mechanical and Artificial Intelligence laboratory (MAIL), at Carnegie Mellon University is broadly interested in the application of machine learning, data science, and molecular dynamics simulations to health and bio-engineering problems. The lab is inherently a multidisciplinary group bringing together researchers with different backgrounds and interests, including mechanical, computer science, bio-engineering, physics, material, and chemical engineering. The mission is to bring the state-of-the-art machine learning algorithm to mechanical engineering. Traditional mechanical engineering paradigms use only physics-based rules and principles to model the world, which does not include the intrinsic noise/stochastic nature of the system. To this end, the lab is developing the algorithms that can infer, learn, and predict the mechanical systems based on data. These data-driven models incorporate the physics into learning algorithms to build more accurate predictive models. They use multi-scale simulation (CFD, MD, DFT) to generate the data.
\end{IEEEbiography}

\EOD

\end{document}